\documentclass[sigconf]{acmart}
\usepackage{balance}

\AtBeginDocument{%
  }

\copyrightyear{2023}
\acmYear{2023}
\setcopyright{acmlicensed}\acmConference[MM '23]{Proceedings of the 31st ACM International Conference on Multimedia}{October 29-November 3, 2023}{Ottawa, ON, Canada}
\acmBooktitle{Proceedings of the 31st ACM International Conference on Multimedia (MM '23), October 29-November 3, 2023, Ottawa, ON, Canada}
\acmPrice{15.00}
\acmDOI{10.1145/3581783.3611936}
\acmISBN{979-8-4007-0108-5/23/10}

\acmSubmissionID{1082}

\begin{document}

\title{Layout Sequence Prediction From Noisy Mobile Modality} 

\author{Haichao Zhang}
\email{zhang.haich@northeastern.edu}
\orcid{0000-0001-9645-3255}
\affiliation{%
  \institution{
  Northeastern University}
  \streetaddress{360 Huntington Ave.}
  \city{Boston}
  \state{Massachusetts}
  \country{USA}
  \postcode{02115}
}

\author{Yi Xu}
\email{xu.yi@northeastern.edu}
\orcid{0000-0001-5857-4179}
\affiliation{%
  \institution{Northeastern University}
  \streetaddress{360 Huntington Ave.}
  \city{Boston}
  \state{Massachusetts}
  \country{USA}
  \postcode{02115}
}

\author{Hongsheng Lu}
\email{hongsheng.lu@toyota.com}
\orcid{0000-0001-9916-1899}
\affiliation{%
  \institution{
Toyota Motor North America}
  \streetaddress{465 N Bernardo Ave}
  \city{Mountain View}
  \state{California}
  \country{USA}
  \postcode{94043}
}

\author{Takayuki Shimizu}
\email{takayuki.shimizu@toyota.com}
\orcid{0000-0002-1068-8510}
\affiliation{%
  \institution{
Toyota Motor North America}
  \streetaddress{465 N Bernardo Ave}
  \city{Mountain View}
  \state{California}
  \country{USA}
  \postcode{94043}
}

\author{Yun Fu}
\email{yunfu@ece.neu.edu}
\orcid{0000-0002-5098-2853}
\affiliation{%
  \institution{Northeastern University}
  \streetaddress{360 Huntington Ave.}
  \city{Boston}
  \state{Massachusetts}
  \country{USA}
  \postcode{02115}
}

\renewcommand{\shortauthors}{Haichao Zhang, Yi Xu, Hongsheng Lu, Takayuki Shimizu, \& Yun Fu}

\begin{teaserfigure}
\centering
  \includegraphics[width=0.95\linewidth]{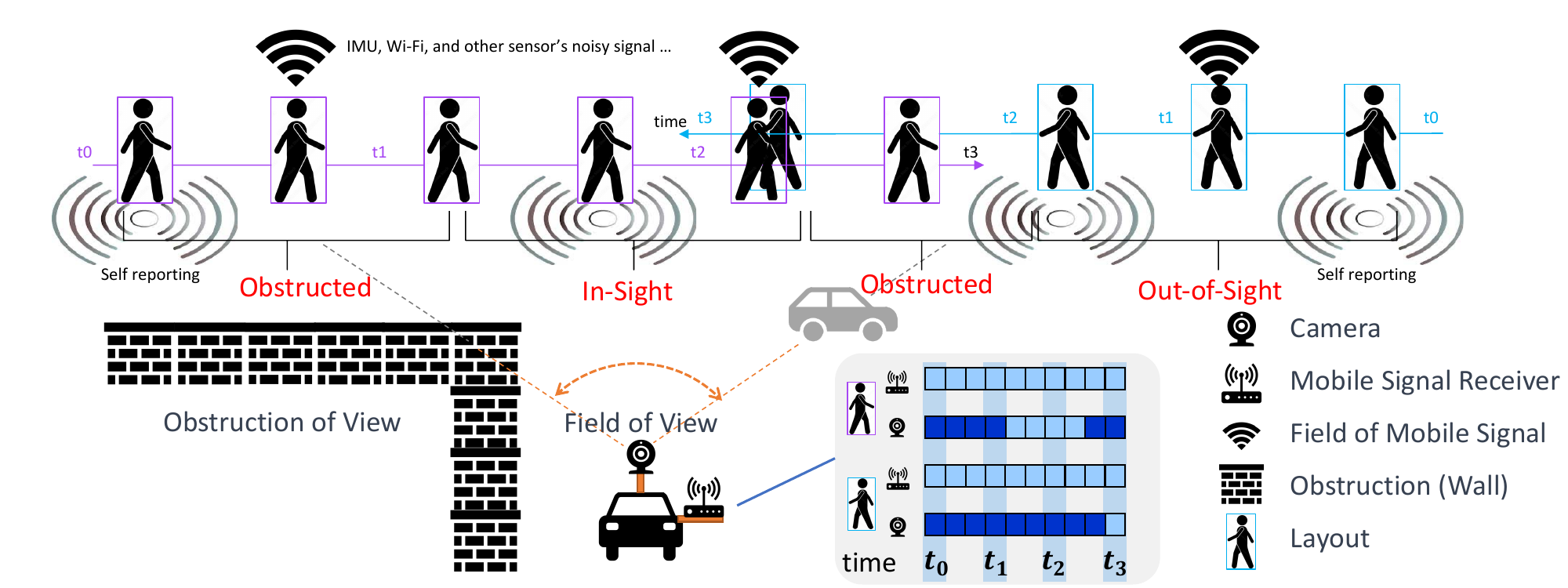}
  \caption{
  Illustration of a typical real-world scenario where a camera-equipped car captures the visual layout sequences and wireless mobile signals from pedestrians and cars. The orange line represents the camera's field of view, while the wireless icon indicates the mobile signal coverage. The figure demonstrates how obstacles, such as walls and cars, can obstruct the view of the camera and disrupt the mobile signals. For example, the pedestrian in the purple bounding box is obstructed by walls and a car before being captured by the camera, and the pedestrian in the blue bounding box walks out of the camera's view and is obstructed by a car. Meanwhile, their mobile sensor's signal remains within the field of reception during this process. The bottom right figure in grey area shows the "visibility" of the pedestrians in the black car's view of the visual and mobile modalities, in which dark color means invisible and light color means visible.
  }
  \label{fig:teaser}
\end{teaserfigure}

\begin{abstract}

Trajectory prediction plays a vital role in understanding pedestrian movement for applications such as autonomous driving and robotics. Current trajectory prediction models depend on long, complete, and accurately observed sequences from visual modalities. Nevertheless, real-world situations often involve obstructed cameras, missed objects, or objects out of sight due to environmental factors, leading to incomplete or noisy trajectories. To overcome these limitations, we propose \textbf{LTrajDiff}, a novel approach that treats objects obstructed or out of sight as equally important as those with fully visible trajectories. LTrajDiff utilizes sensor data from mobile phones to surmount out-of-sight constraints, albeit introducing new challenges such as modality fusion, noisy data, and the absence of spatial layout and object size information. We employ a denoising diffusion model to predict precise layout sequences from noisy mobile data using a coarse-to-fine diffusion strategy, incorporating the Random Mask Strategy, Siamese Masked Encoding Module, and Modality Fusion Module. Our model predicts layout sequences by implicitly inferring object size and projection status from a single reference timestamp or significantly obstructed sequences. Achieving state-of-the-art results in randomly obstructed experiments, our model outperforms other baselines in extremely short input experiments, illustrating the effectiveness of leveraging noisy mobile data for layout sequence prediction. In summary, our approach offers a promising solution to the challenges faced by layout sequence and trajectory prediction models in real-world settings, paving the way for utilizing sensor data from mobile phones to accurately predict pedestrian bounding box trajectories. To the best of our knowledge, this is the first work that addresses severely obstructed and extremely short layout sequences by combining vision with noisy mobile modality, making it the pioneering work in the field of layout sequence trajectory prediction.

\end{abstract}






\maketitle


\section{Introduction}


Trajectory prediction is a critical problem in numerous research fields, including computer vision, robotics, mobile computing, and autonomous driving. The ability to predict future trajectories has the potential to significantly enhance the capacity of these fields to learn human behavior, make informed decisions for vehicles, and pedestrian collision detection. 
However, due to the intricate nature of application scenarios, this has remained a long-standing challenging problem.
Although many works have greatly advanced this field, existing works often assume that the input signals are uncovered, accurate, long, and complete trajectories captured by cameras and estimated by computer vision algorithms, which is too strong an underlying assumption in practice.


Due to the prosperity of computer vision research, numerous topics have attempted to solve this problem and have achieved plausible performances, such as trajectory prediction, object tracking, and video frame prediction. These works have achieved great success in predicting accurate trajectories due to the high accuracy of computer vision algorithms. However, these works share an underlying assumption that relies heavily on the camera to capture the input data. 
Cameras have limited observation sight, which leads to difficulty in blind spots, cover issues, and out-of-sight problems. These limitations greatly impact real-world applications, for example, shortening the reaction time of autonomous driving algorithms and increasing the cost of deploying multi-camera systems for larger sight. Therefore, it is important to explore alternative approaches that can overcome these limitations and produce reliable results in real-world scenarios.

However, less attention pays to the mobile modality that has no out-of-sight problem to estimate the location and motion status of objects without concerns about out-of-view problems. With the development of mobile computing, sensors are now ubiquitous, including IMU mobile sensors~\cite{lu2019imu, han2021hybrid, nirmal2016noise, barrau2020mathematical}, wireless signals~\cite{musarra2019detection, wang2022application}, GPS~\cite{giustiniano2015deep, abbas2019wideep}, satellite data, and user reports to provide out-of-view \cite{omidshafiei2021time, barthelmes2022impact} and blind area\cite{ehlgen2008eliminating, li2010easy} trajectory information.
These sensors have no out-of-sight problem and can be received through a long range without showing in the camera, which is much more ideal than the camera data in the visual modality.
Moreover, computer vision algorithms also face challenges in dealing with occlusions and object appearance changes, which can lead to inaccurate bounding box predictions. In contrast, using sensor data to estimate object location and motion status can provide a complementary approach to improve the accuracy of bounding box prediction.

Despite these beneficial characteristics, there are two main challenges in leveraging mobile modality. The first is that the mobile modality often introduces noise~\cite{giustiniano2015deep, abbas2019wideep, mao1999noise, amiri2007assessment} that makes it difficult to extract accurate signals, particularly when compared to computer vision methods~\cite{morar2020comprehensive}. Another challenge is that important information, such as object size and other detailed information contained in the bounding box, is often missing in these signals, increasing the difficulty of making decisions for high-level tasks like scene understanding. Nevertheless, exploring the use of mobile modality alongside computer vision methods could lead to more accurate and reliable predictions of layout sequence in real-world scenarios. By combining the strengths of both modalities, we can potentially overcome the limitations of each and achieve more robust and efficient solutions.

To address these challenges, we propose a novel approach that combines the strengths of both visual and mobile modalities, aiming to predict a long sequence of layout points from a noisy mobile sensor's data sequence while conditioning on an extremely short or randomly obstructed input visual location or trajectory, such as a single starting point, a few initial observations, or randomly obstructed sequences. By merging these two modalities, we can harness the high accuracy of computer vision tasks and the accessibility and low computational resource consumption of mobile computing. Our approach utilizes a denoising diffusion model to denoise the mobile modality and transform it into layout sequences.

In the denoising diffusion model (LTrajDiff), we first propose a Random Mask Strategy (RMS in Section~\ref{rms}) to simulate the random obstruction input signals. Then, we propose a Siamese Masked Encoding Module (Section~\ref{siamese}) to extract the 3D layout information from limited observations with a Layout Extracting Module (LEM in Section~\ref{lem}) and align the inputs of mobile and visual modalities while extracting temporal information with a Temporal Alignment Module (TAM in Section~\ref{tam}). Finally, to better leverage the visual and mobile modalities, we propose a Modality Fusion Module (MFM in Section~\ref{mfm}) to merge the two modalities into the diffusion model. Our method outperforms all other baselines in both extremely short observations and randomly obstructed observations, and we conduct thorough ablation studies to illustrate the function of each component module.


The contributions of this paper are as follows:
\begin{itemize}
\item We introduce a novel task to predict layout sequence trajectories that combine the benefits of visual and mobile modalities to overcome their limitations and improve sequence prediction accuracy from noisy mobile modality and extremely short or randomly obstructed observations from the visual modality.
\item We propose a denoising diffusion model(LTrajDiff) that accurately predicts trajectory sequences from noisy data and short or randomly obstructed layout sequences, reducing the difficulty of predicting long sequences and improving prediction accuracy.

\item We propose a RMS module to simulate and adapt to obstruction observations, an MFM module, and a Siamese Masked Encoding Module consisting of TAM and LEM to extract, leverage, and merge information from visual and mobile modalities.

\item We demonstrate significant improvement over existing methods, conducting extensive ablation studies to prove the efficiency of our model, indicating that our approach can be a valuable tool for predicting layout sequences from noisy data and short or severely obstructed input sequences in neural network-based applications.
\end{itemize}

\section{Related Works}
The problem of layout sequence prediction is situated at the intersection of trajectory prediction and vision-wireless fusion. In trajectory prediction, the goal is to forecast future object positions based on observations. Vision-wireless fusion, on the other hand, seeks to integrate mobile modalities such as mobile computing and wireless signals with computer vision to tackle visual tasks. 

\subsection{Vision-Wireless Fusion}
Vision-Wireless Fusion aims to leverage mobile modalities to address challenges in computer vision. Previous work has demonstrated the feasibility of integrating mobile computing and wireless signals with computer vision. For example, Liu et al.\cite{liu2020vision} employed wireless trajectories to tackle challenges in person re-identification, such as variations in pedestrian postures and clothing. Papaioannou et al.\cite{papaioannou2015accurate} used radio and magnetic maps to address camera instability and changes in field of view when tracking people in dynamic industrial settings. Additionally, Alahi et al.~\cite{alahi2015rgb} enriched the RGB-D camera modality with mobile modalities to improve localization accuracy by estimating distance using wireless signals from individuals' cell phones. Although these works successfully integrated mobile modalities into computer vision tasks, they have received less attention in solving the problem of layout sequence prediction. In this work, we aim to explore the potential of vision-wireless fusion to address the challenges of layout sequence prediction.

\subsection{Trajectory Prediction}

Our work is the first to focus on layout sequence prediction, and while there are no prior works directly related to our task, trajectory prediction is a closely related topic. The objective of trajectory prediction is to forecast the future positions of objects based on a sequence of observations. Trajectory prediction can be categorized into two types: those based on wireless mobile computing and those based on computer vision.

\subsubsection{Trajectory Prediction Based on Wireless Sensors}

Wireless mobile computing-based trajectory prediction uses only data from wireless sensors as input and estimates the relative locations of mobile devices using the noisy sensor data. For example, Liu et al.\cite{liu1998mobility} use Kalman filtering techniques to estimate the movement behavior of mobile users for trajectory prediction, which helps manage the rerouting and mobile handoff in wireless ATM networks management. Mlika et al.\cite{mlika2023user} propose to use quantum reservoir computing to predict mobile users' trajectories in mobile wireless networks to address mobility management problems in self-organizing and autonomous 6G networks. Gao et al.~\cite{9350660} propose a method of fingerprint location for Wi-Fi signals assisted by smartphone built-in sensors to improve the location accuracy and reduce the adverse effects of environmental factors.
However, these trajectory prediction methods purely based on mobile modality suffer from the limitations of the sensors, which results in inaccurate and limited information. The strength of wireless signals is used to estimate the distance between the receiver and communication devices like smartphones, and IMU or other sensors capture the movement or speed information of users, or GPS locations are used, which cannot be used in outdoor locations. Therefore, the acquired information is often too noisy and inaccurate.

\subsubsection{Trajectory Prediction Based on Computer Vision}

In contrast, computer vision-based trajectory prediction methods analyze images, videos, graphs, point clouds, and other media to provide accurate locations. For example, Li et al.\cite{Li_2022_CVPR} use graph-based transformers to model historical trajectory with a memory graph to smooth and predict multiple trajectories. Liang et al.\cite{liang2019peeking} predict pedestrian future paths/
trajectories from videos by modeling their interaction between human behaviors and surroundings.
Qian et al.\cite{qian2022trajectory} propose a method in an "easy-to-hard" schema to hierarchically divide the model into inter-trajectory view and intra-trajectory level.
Ye et al.\cite{Ye_2021_CVPR} propose a temporal point cloud learning network to capture spatial and temporal information and perform trajectories prediction directly from point cloud.
However, computer vision-based trajectory prediction methods are limited by restricted scenarios, limited field of view, and obstruction, which greatly limit real-world applications, and many works are confined to indoor scenarios with a good viewpoint that can see every object, such as sports events.

\subsubsection{Layout Sequences Prediction Based on Combination of Visual and Mobile Modality}



To overcome the limitations of the wireless sensor-based trajectory prediction methods and the problems of restricted scenarios, limited field of view, and obstruction in computer vision-based trajectory prediction, we propose to combine mobile and visual modalities to solve the layout sequence prediction problem. We do not directly work on trajectory prediction because layout sequence prediction requires object size, spatial information, and projection status, which are more challenging but more meaningful to solve. Our novel approach aims to fuse the strengths of both visual and mobile modalities to enhance the accuracy and reliability of layout sequence predictions in complex real-world scenarios. By leveraging the high accuracy of computer vision tasks and the accessibility and low computational resource consumption of mobile computing, we hope to pave the way for more advanced solutions in the field of layout sequence prediction.
\begin{figure*}[ht]
  \centering
  \includegraphics[width=0.95\linewidth]{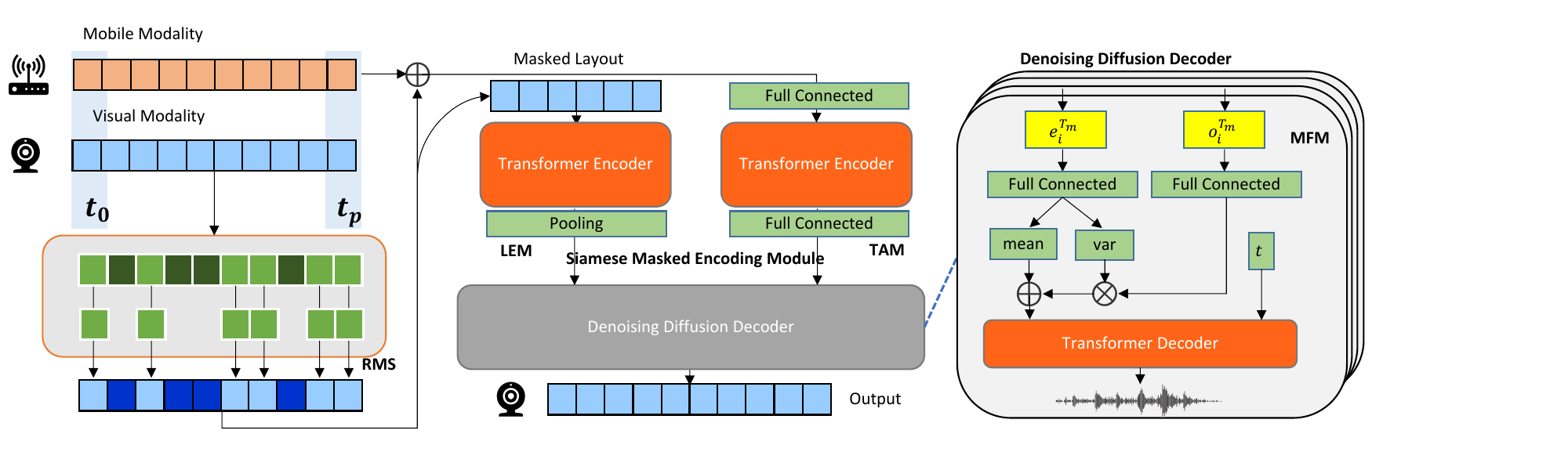}
  \caption{
  Overview of Our Denoising Diffusion Model (LTrajDiff). Our model takes sensor data from a mobile modality and a few timestamps of layout sequence as input, and outputs imputed 3D layout sequences in the form of projected layout information with estimated bounding box size and depth. The model comprises a diffusion model and a Siamese Masked Encoding Module to denoise noisy mobile data and predict accurate layout sequences. Leveraging object size inference, our model can generate accurate layout sequences from incomplete and noisy data, making it a promising solution for pedestrian movement prediction in real-world scenarios.
  }
  \label{fig:arch}
  
\end{figure*}
\section{Problem Definition}
We consider a set of $N$ agents denoted by $\Omega = {1, 2, ..., N}$, observed between timestamps $t_1$ and $t_p$. The coordinate of the $i$-th agent at timestamp $t_q$ is denoted by $x_i^{t_q}$. The set of coordinates between timestamps $[t_q:t_p]$ is $X_i^{t_q:t_p} = \{ x^q_i, ..., x^t_i, ..., x^p_i \}$, where $x_t \in \mathbb{R}^2$. The layout sequence of the $i$-th agent at timestamp $t$ is denoted by $l_i^t \in \mathbb{R}^5$, and the mobile modality signal of the $i$-th agent at timestamp $t$ is denoted by $s_i^t \in \mathbb{R}^{19}$. The layout sequence $l_i^t$ consists of a bounding box $b_i^t \in \mathbb{R}^4$ and a depth value $d_i^t \in \mathbb{R}^1$. The mobile modality signal includes IMU sensor data sequences and Wi-Fi sensor data sequences.
To simulate missing observations due to obstructions or out of sight of cameras, we define a mask matrix $M_i^{t_q:t_p} = \{ m^q_i, ..., m^t_i, ..., m^p_i \}$, where $m_i^t \in {0,1}$ represents whether the observation of the $i$-th agent at timestamp $t$ is missing or not.
Our task is to predict the visual layout sequence trajectory $\hat{L}_i^{t_q:t_p}$ in the camera view, given noisy mobile sensor data $S_i^{t_q:t_p}$ and a randomly obstructed and missing short visual observation. We simulate the visual observation by masking the ground truth visual sequence $L_i^{t_q:t_p}$ with the missing mask matrix $M_i^{t_q:t_p}$, which yields the input visual sequence $I_i^{t_q:t_p}$:
\begin{equation}
I_i^{t_q:t_p} = L_i^{t_q:t_p} \bullet  M_i^{t_q:t_p}.
\end{equation}

Formally, our goal is to learn a function $\textit{f}(\cdot)$ that can predict $\hat{L}_i^{t_q:t_p}$ from $I_i^{t_q:t_p}$ and $S_i^{t_q:t_p}$:
\begin{equation}
\hat{L}_i^{t_q:t_p} = \textit{f}(I_i^{t_q:t_p}, S_i^{t_q:t_p}) .
\end{equation}
Our approach is based on a denoising diffusion model with a masked encoding module that denoises the noisy mobile data and infers object size to predict accurate bounding box trajectories.

\section{Methodology}
The overall architecture of our proposed method is illustrated in Fig.~\ref{fig:arch}. Our method addresses the task of predicting layout sequences by embedding noisy mobile data and incomplete layout sequences with a unified framework that predicts layout sequences.


\subsection{Random Mask Strategy (RMS)}
\label{rms}

First, we use a random mask model to simulate the obstructed and out-of-sight processes in real-world scenarios. We improve on the approach used in MAE~\cite{he2022masked} by randomizing not only the position of the mask but also the ratio of masked elements to the overall length of the input sequence. We define the mask model as follows:

\begin{equation}
M_i^{t_q:t_p} = \mathbf{Shuffle} ([0]_{(q-p)*r} \circ [1]_{(q-p)*(1-r)}) \\
, r \sim U(0,1) ,
\end{equation}

where $\mathbf{Shuffle}$ is a function that rearranges the indices of the mask matrix, $r$ denotes the random mask ratio, and $U(0,1)$ is a uniform distribution between 0 and 1. $[a]_{b}$ denotes a list with b elements of value a, and $\circ$ denotes the concatenation operation between two lists.


\subsection{Siamese Masked Encoding Module}
\label{siamese}
In this section, we aim to address the two challenges presented in the introduction. Firstly, in Section~\ref{tam}, we tackle the issue of noise that is often introduced in mobile modality, which makes it difficult to extract accurate signals. Secondly, in Section~\ref{lem}, we handle the challenge of missing important information, such as object size and other details contained in bounding boxes, in these trajectory's signals.

Although extracting information from noisy mobile modality and predicting layout information from a limited number of timestamps is challenging, real-world scenarios demand that our model can handle both tasks simultaneously. To achieve this, we propose a pair of Siamese encoders comprising a temporal alignment module and a layout extraction module. These encoders aim to abstract the layout information and target object size and layout information.

Aiming to synchronize and align the noisy data and incomplete layout sequences, we employ a temporal alignment module to encode the concatenated feature of incomplete visual modality and noisy wireless sensor signals in the mobile modality. Meanwhile, we use a layout extraction module to encode the layout under the masked area.

\subsubsection{Temporal Alignment Module (TAM)}
\label{tam}
The temporal alignment module is a transformer~\cite{vaswani2017attention} module with fully connected layers before and after. It takes the incomplete visual layout sequence $I_i^{t_q:t_p}$ and the noisy wireless sensor signals $S_i^{t_q:t_p}$ as inputs and outputs the aligned feature embedding $E_i^{t_p:t_q}$.

The transformer module consists of multiple layers of self-attention and feed-forward neural networks. Each layer first applies a self-attention mechanism to capture the temporal dependencies between different timestamps. Then, a feed-forward neural network is used to transform the attention output, followed by a residual connection and a layer normalization operation. This process is repeated for multiple layers, and the output of the last layer is used as the temporal alignment feature embedding $e_\mathbf{TAM}^{t_p:t_q}$. This process can be denoted as,
\begin{equation}
e_i^{t_p:t_q} = \mathbb{E}_\mathbf{TAM}((L_i^{t_q:t_p} \bullet M_i^{t_q:t_p} ) \oplus S_i^{t_p:t_q}),
\end{equation}
where $\oplus$ denotes the concatenation operation, $\hat{L}_i^{t_q:t_p}$ is the layout sequence with missing data imputed using interpolation or extrapolation methods, and $\mathbb{E}_\mathbf{TAM}$ is the temporal alignment module.

\subsubsection{Layout Extracting Module (LEM)}
\label{lem}

The layout extracting module aims to infer the object size, layout, and other detailed information, such as projection orientation, contained in the bounding box. To better model the layout information, the LEM takes the masked area of the visual modality as input. We use a transformer-based encoder to process the masked area, but since the masked area has a random length due to the random masked ratio, the length of the embedded feature is not fixed. To address this issue, we use a pooling layer to map the results into fixed lengths for the subsequent module to process.
Formally, we denote the timestamps set $T_m$ that are not masked as follows:
\begin{equation}
T_m = \{t \mid t_q \leq t \leq t_p, m_i^{t} = 1\},
\end{equation}
where the right-hand side of the notation $\leq$ represents the condition for the timestamps set $t$.
Then, the masked area of the visual modality $\dot{I}i^{t_q:t_p}$ can be defined as follows:
\begin{equation}
\dot{I}_i^{t_q:t_p}= I_i^{T_m},
\end{equation}

here, $I_i^{T_m}$ is first processed by the transformer-based encoder and then fed into a pooling layer. Specifically, we apply a mean calculation on the timestamp dimension. Consequently, the embedded layout feature $o_i^{T_m}$ from LEM can be computed as follows:

\begin{equation}
o_i^{T_m} = \frac{1}{T}\sum_{t=1}^T \mathbb{E}_\mathbf{LEM}( I_i^{T_m} ),
\end{equation}

where $\mathbb{E}_\mathbf{LEM}$ is the layout extracting module, and $T$ is the length of the set $T_m$. This formula computes the mean of the feature $\mathbb{E}_\mathbf{LEM}( I_i^{T_m} )$ over the time dimension.

By combining the temporal alignment module (TAM) and the layout extracting module (LEM), our Siamese Masked Encoding Module can effectively synchronize and align the noisy data and incomplete layout sequences and infer the object size, layout, and other detailed information contained in the bounding box. In doing so, our model can handle the challenges of extracting information from noisy mobile modality and predicting layout information from a limited number of timestamps in complex real-world scenarios.

\subsection{Denoising Diffusion Decoder}
In this section, we implement a decoder based on the diffusion model to decode the features of layout representation and temporal alignment representation. The diffusion model is a powerful generative model capable of handling a wide range of data distributions and generating high-quality samples. We use a coarse-to-fine diffusion model to gradually remove noise and synthesize the predicted layout sequence. In particular, the diffusion model allows us to iteratively refine the noisy input data and generate a denoised output. 
We adopt a diffusion model to jointly obtain embeddings from the layout feature and the temporal alignment feature.

\subsubsection{Modality Fusion Module (MFM)}
\label{mfm}

Let $e_i^{t_p:t_q}$ denote the temporal alignment feature embedding for node $i$ from time step $t_p$ to time step $t_q$, and let $o_i^{T_m}$ denote the embedded layout feature for node $i$ at time step $T_m$.
First, we use the fully-connected layer $f$ to map the temporal alignment feature embedding $e_i^{t_p:t_q}$ to the mean $\mu$ and variance $\sigma$ of the normal distribution:
\begin{equation}
\mu,\sigma = f(e_i^{t_p:t_q}).
\end{equation}
Next, we use the fully-connected layer $g$ to map the embedded layout feature $o_i^{T_m}$ to a modality fusion feature $z_i$:
\begin{equation}
z_i = \sigma \bullet g(o_i^{T_m}) + \mu.
\end{equation}

\subsubsection{Diffusion Model}
The diffusion model is implemented with a transformer decoder. In the $k$-th iteration, the transformer decoder takes the modality fusion feature $z_i$, time embedding $t$, and a noisy layout sequence by adding noise in the forward process $\epsilon$ from $L_i^{t_q:t_p}$ to predict the noise item $\hat{\epsilon}$ in the backward process in the diffusion model. The forward process of the diffusion model is,
\begin{equation}
y_k = \sqrt{a_k} y_0 + \sqrt{1-a_k} \epsilon,
\end{equation}
where $y_k$ is the $k$-th output of the diffusion model, so $y_0 = L_i^{t_q:t_p}$.
$\epsilon = \mathcal{N}(0,1)$. $a_t = 1-\beta_t$, where $a_t$ is calculated from a variance schedule ${ \beta_t \in (0,1) }_{t=1}^{T}$ which is an equivariant series that controls the step sizes of the diffusion model.
The reverse process of the diffusion model can be denoted as,
\begin{equation}
\hat{\epsilon} = \mathcal{F} (z_i, t, y_{k-1}),
\end{equation}
in which $\mathcal{F}$ is the transformer decoder. So, the loss function will be,
\begin{equation}
\mathcal{L} = \lambda \mathcal{L}_2( \hat{\epsilon} - \epsilon),
\end{equation}
in which the $\mathcal{L}_2$ is the L2 loss function, and $ \lambda$ is the weighting coefficients.

\subsubsection{Training}
The denoising diffusion decoder is trained from scratch through iterative steps. The model performs a single iteration of the denoising diffusion process in each training step. During training, we sample a random diffuse step between [0, K], where K represents the total number of iterations. In each iteration, the model aims to predict the noise introduced in that diffusion step to reveal the potential signal for the next denoising step. During testing, the model is initially fed with a randomly sampled noise from a normal distribution in the first iteration. The predicted denoised vector from each iteration is then added to the predicted results of the previous iteration in the subsequent iteration. This iterative process is repeated K times until we obtain the final results.

\subsection{Implementation Details}
In this section, we provide detailed information about the implementation of our method. We use the Adam \cite{KingmaB14} optimizer to jointly train the entire model. The learning rate is set to 1e-3. An exponential learning rate decay scheduling is applied, with the learning rate decaying exponentially by multiplying a gamma value of 0.98. We use two NVIDIA RTX 2080Ti GPUs with 11 GB CUDA memory for training. The batch size for training is set to 256.

\section{Experiments}

\subsection{Datasets}
\label{dataset}
We evaluate our method on two datasets, the Vi-Fi Multi-modal Dataset~\cite{liu2022vi,liu2021lost} and the H3D Dataset~\cite{360LiDARTracking_ICRA_2019}, for the problem of layout sequence prediction from noisy mobile data. To assess our proposed method, we require datasets that contain layout sequences (trajectories, bounding boxes, depths, etc.) in the vision modality and wireless sensor signals (IMU, GPS, and other sensor signals) in the mobile modality. Please refer to the details of the datasets in Appendix.~\ref{app:dataset}.

\subsection{Metrics}
Since our work is the first to focus on the task of layout sequence (trajectories, bounding box, depth) prediction, we cannot directly use the metrics from trajectory prediction tasks. We need new metrics to better evaluate our task. Please note that our layout sequences are in the form of concatenated left-bottom coordinates and width, height, and depth of the bounding box.

\subsubsection{Mean Square Error per Timestamp (MSE-T)}
We calculate the Mean Square Error between two layout sequences, ignoring spatial relationships for simplicity. The result is an average over the timestamp dimension. The formula can be denoted as:

\begin{equation}
\mathcal{L}{MSE-T} = \frac{1}{T}\sum_{t=1}^T {\left| L_i^{t_q:t_p} - \hat{L_i}^{t_q:t_p} \right|}^2     ,
\end{equation}

here, $\mathcal{L}_{MSE-T}$ is the calculated MSE-T metric. $T$ is the length between $t_q:t_p$. $ L_i^{t_q:t_p} $ and $\hat{L}_i^{t_q:t_p}$ are the visual layout sequence trajectory and its ground truth.

\subsubsection{Intersection over Union with Depth (IoU-D)}
Because the predicted layout sequence contains depth, coordinates, width, and height simultaneously, applying MSE directly ignores the spatial information between these elements. We modify the standard IoU to include depth information and better leverage width and height for our task.

\begin{equation}
\mathcal{L}_{IoU-D} = \frac{(l_1 * l_2) }{w * h + \hat{w} * \hat{h} - l_1 * l_2} * \frac{|d - \hat{d}|} {max(d, \hat{d})}   ,
\end{equation}

here, $d$ and $\hat{d}$, $w$ and $\hat{w}$, and $h$ and $\hat{h}$ are the depth, width, and height in the predicted layout sequence and ground truth, respectively. $l_1$ and $l_2$ are the overlap areas of two bounding boxes on width and height.



\subsection{Baselines}
\label{sec:exp:baselines}
Predicting layout sequences from noisy mobile modality is a new task, so we implement four baselines with different input settings for evaluation. Note that the input for these baselines is a concatenation of the incomplete visual layout sequence and the noisy wireless sensor signals, while the output is complete visual layout sequences. As our proposed method is a diffusion-based model with only $\mathcal{L}_2$ loss applied, these baselines use $\mathcal{L}_2$ loss during training.

\subsubsection{ViTag~\cite{cao2022vitag}}
Vi-Tag is the current state-of-the-art method on the Vi-Fi dataset~\cite{liu2022vi,liu2021lost} for the vision-motion identity association task, which aims to match identities in cameras in the vision modality with wireless signals in the mobile modality. The core of Vi-Tag is the X-Translator model, which translates the visual modality to the mobile modality and finds the most similar wireless signals to match the identities in both modalities. We modify the X-Translator to translate the concatenation of the incomplete visual layout sequence and the noisy wireless sensor signals into complete visual layout sequences.

\subsubsection{UNet~\cite{rossi2021human}}
UNet is a well-known auto-encoder backbone architecture widely used in various research topics, including image segmentation, generative models, action recognition, and trajectory prediction. To adapt it to our task, we need to reshape the input and output into 2D images. Despite its excellent performance in these tasks, the main drawback of UNet is that it requires input and output to have a fixed shape and cannot process data with variable lengths.

\subsubsection{Other Seq2seq Models}
We also adapt other sequence-to-sequence models originally designed for trajectory prediction-related tasks, such as LSTM~\cite{shi2018lstm} and Transformer~\cite{giuliari2021transformer}. The models are implemented as encoder-decoder architectures. The inputs are synchronized on the dimension of timestamps before being fed into the seq2seq model.

\begin{figure}[tb]
  \centering
  \includegraphics[width=0.95\linewidth]{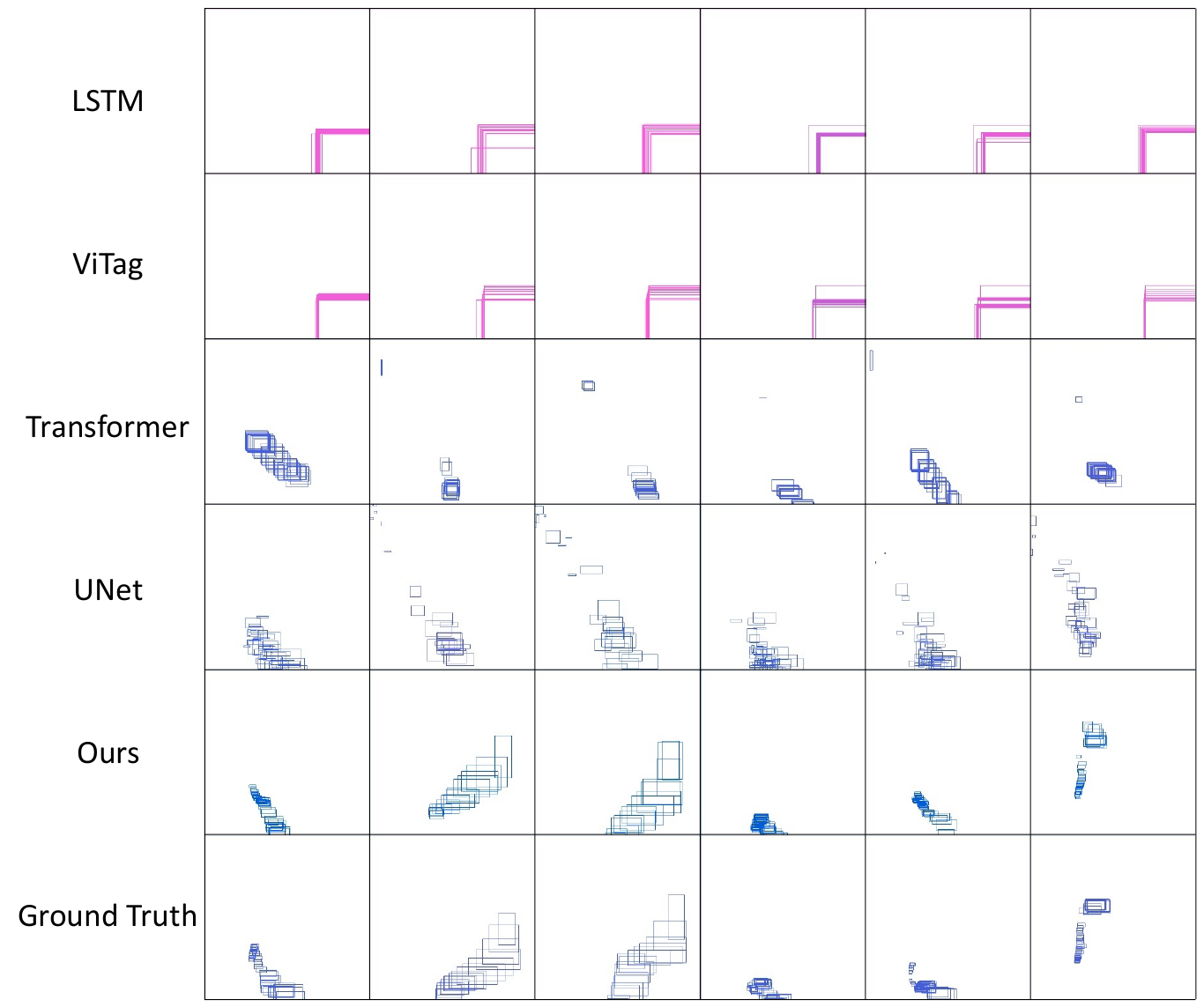}
  \caption{
Comparison of Visualization Results. The figure above provides a visual comparison of the predicted layout sequences generated by different methods. Each method's predicted layout sequence is presented, and the corresponding objects are depicted as rectangles representing their bounding boxes. The color of each bounding box reflects the depth of the object, with darker blue shades indicating closer proximity. It is worth noting that in the first and second rows, the LSTM and ViTag methods encounter difficulties in converging on this task, as evidenced by their less accurate predictions. For the purpose of comparison, only the outputs are displayed in this figure.
  }
  \label{fig:visualization}
\end{figure}

\subsection{Quantitative Comparison Experiments (randomly obstructed experiments)}

In this section, we present the quantitative experiments, with details of the implemented baselines listed in Section ~\ref{sec:exp:baselines}.


As shown in Table ~\ref{exp:compareall}, under the random visibility mask with a random ratio setting, the results on the MSE-T dataset are relatively high. The H3D dataset shows better overall MSE-T scores because it has more mobile modality sensors. This provides more data sources and dimensions that can partly eliminate the noise in the mobile modality, helping the model learn better movement cues to recover the movement status and ultimately lead to better layout sequences. Among the baselines, except for our model yielding the best results, the UNet~\cite{rossi2021human} and Transformer~\cite{giuliari2021transformer} also perform well in terms of the MSE-T metric on the Vi-Fi dataset. However, their IoU-D results are worse than ours, as are the MSE-T results on the H3D dataset.

\begin{table}[t]
\begin{center}
\begin{tabular}{|l|c|c|c|}
\hline
 Dataset &   H3D~\cite{360LiDARTracking_ICRA_2019}   & \multicolumn{2}{|c|}{Vi-Fi~\cite{liu2022vi}} \\
Metrics &   MSE-T $\downarrow$ & MSE-T $\downarrow$  & IoU-D $\uparrow$ \\
\hline \hline
LSTM~\cite{shi2018lstm}  & 452.14 & 432.33 & 0.04 \\ 
ViTag~\cite{cao2022vitag} & 455.61 & 421.42 & 0.04\\   
Transformer~\cite{giuliari2021transformer} & 5.17 & 58.29 & 0.42\\    
UNet \cite{rossi2021human}  & 5.92 & 58.79 & 0.43 \\  
MID  \cite{gu2022stochastic} & 13.48  & 64.07 & 0.18 \\
HIVT \cite{zhou2022hivt}     & 2.88 & 66.07 & 0.19 \\
\hline
LTrajDiff (Ours) & 2.72 & 56.13 & 0.69  \\
\hline
\end{tabular}
\end{center}
\caption{\textit{Evaluation of the Denoising Diffusion Model against Baseline Approaches on H3D and Vi-Fi Datasets with Two Metrics.} Compared to other baselines adapted to random length and random visibility masks, our proposed model achieves superior performance in terms of MSE-T and IoU-D. Note that the layout sequences in the H3D dataset are not bounding boxes (objects with mobile signals are not annotated with a bounding box), so we can only apply MSE-T for evaluation.}
\label{exp:compareall}
\end{table}
\subsection{Extremely Short Input Experiments}


As shown in Table ~\ref{exp:compare:extreshort}, our model outperforms the baselines in the extremely short input experiments without the Phase-I and Phase-II training schemes. The second-best baseline, UNet, also achieves good results in this experiment, while LSTM and Vi-Tag do not. This experiment showcases the ability to predict with extremely short inputs.

\begin{table}[tb]
\begin{center}
\begin{tabular}{|l|c|c|}
\hline
Model & Phase I $\downarrow$ & Phase II $\downarrow$ \\
\hline\hline
LSTM~\cite{shi2018lstm} & 110.11 & 116.24 \\
ViTag~\cite{cao2022vitag} & 110.30 & 110.32 \\
Transformer~\cite{giuliari2021transformer} & 28.28 & 28.27 \\
UNet \cite{rossi2021human} & 3.42 & 5.52 \\
HIVT \cite{zhou2022hivt}     & - & 17.51  \\
MID  \cite{gu2022stochastic} & -  & 13.32  \\

\hline
LTrajDiff(Ours) & - & 4.48 \\
\hline
\end{tabular}
\end{center}
\caption{\textit{Evaluation on Extremely Short Inputs of Our Model Against Baseline Approaches.} Phase-I indicates when input has 10/50 of the predicted length, while Phase-II indicates the input has 1/50 of the predicted length. Compared to other baselines that require a Phase-I and Phase-II training strategy to adapt to extremely short input sequences (only one timestamp), our proposed model achieves superior performance in terms of MSE-T. Note that the $\uparrow$ means high is better and the $\downarrow$ means low is better.}
\label{exp:compare:extreshort}
\end{table}

\subsection{Visualization Results}
The visualization results are shown in Fig.\ref{fig:visualization}.
Upon observing the results, it becomes evident that both ViTag and LSTM encounter difficulties in converging on this task, resulting in inconsistent and inaccurate predicted layout sequences. On the other hand, Transformer and UNet exhibit the ability to approximate the trajectory and depth to some extent, but they struggle with accurately inferring the size of the bounding boxes. In contrast, our proposed model showcases superior performance by accurately predicting a rectangular box trajectory with precise bounding box sizes. This highlights the effectiveness of our method in addressing the challenges associated with layout sequence prediction.

\subsection{Ablation Study of Modules}
In this experiment, we remove each component module in our model to demonstrate the functions of each module. The results of the experiments are shown in Table~\ref{exp:ablation:module}. We will introduce the settings of the experiment and analyze the results in this section.




As shown in Table~\ref{exp:ablation:module}, the performance drops if we remove any of the module elements. Specifically, removing the MFM module causes both metrics to drop slightly, but it remains better than other ablations. This indicates that MFM helps the model better fuse the features from TAM and LEM. Without MFM, the performance drops due to difficulty in modality fusion.
Removing the LEM module causes a slight drop in MSE-T but leads to a significant drop in IoU-D, which indicates that the LEM module has a strong relationship with IoU-D. This means that LEM plays a role in extracting layout spatial information to help the model predict better 3D layouts in bounding boxes and depth.
Without the TAM module, the performance of both MSE-T and IoU-D degrades because it is difficult for the model to make predictions if temporal information is lost. The relationship between LEM and TAM is that TAM helps the model understand spatial information, but without TAM to provide temporal alignment information to the model, both performances degrade. These results demonstrate the efficiency of TAM and reveal the relationship within the Siamese Masked Encoding Module.
The poor performance of "w/o RMS" indicates that the model will have out-of-distribution issues because the inputs during training and testing will be too different for the model to adapt. This result shows that RMS can simulate incomplete testing samples to aid model training.

\begin{table}[tb]
\begin{center}
\begin{tabular}{|c|c|c|}
\hline
\qquad Model Variant  \qquad\qquad & MSE-T $\downarrow$ &  IoU-D $\uparrow$ \qquad\\
\hline\hline
w/o RMS~(\ref{rms} ) & 307.41 & 0.08\\
w/o MFM~(\ref{mfm} ) & 113.55 & 0.45 \\
w/o TAM~(\ref{tam} ) & 295.93 & 0.15\\
w/o LEM~(\ref{lem} ) & 64.07  & 0.18\\
\hline
Complete model       & 56.13  & 0.69\\
\hline
\end{tabular}
\end{center}

\caption{\textit{Ablation Study of Model Components.}
We removed individual modules from our complete model (corresponding subsections in the Method Section are listed in brackets). The results demonstrate that the IoU-D metric experiences a significant drop when the TAM and LEM modules are excluded, while MSE-T decreases substantially when the TAM or MFM module (responsible for processing the features from TAM and LEM) is removed. Without the RMS module, the performance is poor. Note that the $\uparrow$ signifies higher values are better and the $\downarrow$ indicates lower values are better.
}

\label{exp:ablation:module}
\end{table}

\subsection{Ablation Study of Modality}
This experiment aims to determine the contributions of visual and mobile modalities in helping the model predict layout sequences.
\subsubsection{Experiments Setup.}
To isolate the modality, for each modality, we completely remove the incomplete layout sequences (in visual modality) or noisy mobile sensor data sequences (in mobile modality). Although we attempt to keep the architecture unchanged when training with a single modality, the LAM module takes only the unmasked area in incomplete layout sequences as input. As a result, we remove the LAM module when implementing the "w/o Visual Modality", and the TAM module only takes the noisy mobile sensor data sequences (in mobile modality).
In the implementation of "w/o Mobile Modality", we simply remove the noisy mobile sensor data sequences (in mobile modality) from the input of the TAM module.
\subsubsection{Results Analysis.}
As shown in Table.~\ref{exp:ablation:modality}, when either visual or mobile modality is removed from the inputs, the model's performance declines significantly. This indicates that both modalities are crucial for the layout prediction process. Furthermore, the similar drops in performance for both MSE-T and IoU-D metrics suggest that the two modalities, as well as the extracted spatial layout information and temporal alignment information, have comparable levels of importance in layout sequence prediction.

\begin{table}[tb]
\begin{center}
\begin{tabular}{|c|c|c|}
\hline
\qquad Modality Variant  \qquad\qquad & MSE-T $\downarrow$ &  IoU-D $\uparrow$ \qquad\\
\hline\hline
w/o Mobile Modality & 387.86 & 0.29\\
w/o Visual Modality & 362.33 & 0.33\\
\hline
Mobile + Visual Modality       & 56.13  & 0.69\\
\hline
\end{tabular}
\end{center}

\caption{\textit{Ablation Study of Input Modality.} Our model's performance is greatly impacted when either the visual or mobile modality is removed from the input. Results on both MSE-T and IoU-D metrics drop significantly without both modalities as inputs. This experiment confirms the importance of both modalities in the layout prediction process. Note that the $\uparrow$ means high is better and the $\downarrow$ means low is better.}
\label{tab:chart}
\label{exp:ablation:modality}
\end{table}

\section{Limitation Discussion}
The main distinction in computation lies in the inclusion of additional dimensions, such as depth, x and y coordinates, height, and width of the bounding box, compared to traditional trajectory prediction. However, in terms of time cost, there is no significant difference between layout sequence prediction and trajectory prediction. Both tasks involve predicting coordinates and dimensions, and the additional dimensions in layout sequences prediction do not significantly impact the computational complexity.

That being said, while our LTrajDiff achieves remarkable performance, the diffusion model requires forwarding K times to predict the final layout sequences during the testing process, which can be time-consuming. One potential avenue to mitigate this limitation is to explore deep learning acceleration techniques and model pruning methods for the diffusion model, which can significantly reduce the inference time. Additionally, future researchers could investigate alternative models that strike a balance between accuracy and acceleration. Furthermore, our current model provides a projection of the bounding box with depth information. However, there is room for improvement by incorporating real-time volume segmentations or 3D mesh representations to gather more comprehensive information. We will discuss these potential avenues to highlight the future directions of research in our conclusion.

\section{Conclusion}

Our study addresses a prevalent issue in real-world scenarios, where inputs are often assumed to be unobstructed, complete, and accurate camera trajectory sequences. We introduce a novel approach that leverages noisy mobile modality data to consider objects that are out-of-sight or obstructed by complex environments. To overcome the challenges associated with noisy mobile modalities, we propose a denoising diffusion model. Specifically, we introduce the Modality Fusion Module to tackle modality fusion challenges, the Siamese Masked Encoding Module to leverage information from both visual and mobile modalities to address noisy data issues, and the Layout Extracting Module to handle missing spatial layout and object size information. Through experiments, we demonstrate the efficiency and superiority of our architecture and contributions. Our model achieves state-of-the-art performance
This demonstrates the effectiveness of our approach in leveraging noisy mobile data to predict layout sequences. To the best of our knowledge, this is the first work to address severely obstructed and extremely short input layout sequences by leveraging the combination of computer vision and noisy mobile modality. Our findings pave the way for further research and development in the fusion of diverse data sources to tackle real-world challenges.

\section{Acknowledgement}
This research was sponsored by Toyota Motor North America.

\bibliographystyle{ACM-Reference-Format}
\balance
\bibliography{mm23}

\end{document}